RESEARCH ARTICLE

# CurFi: An automated tool to find the best regression analysis model using curve fitting

Ayon Roy | Tausif Al Zubayer | Nafisa Tabassum | Muhammad Nazrul Islam | Md. Abdus Sattar

Department of Computer Science and Engineering, Military Institute of Science and Technology, Dhaka, Bangladesh

**Correspondence**
Muhammad Nazrul Islam, Department of Computer Science and Engineering, Military Institute of Science and Technology, Mirpur Cantonment, Dhaka 1209, Bangladesh.
Email: nazrul@cse.mist.ac.bd

**Abstract**

Regression analysis is a well known quantitative research method that primarily explores the relationship between one or more independent variables and a dependent variable. Conducting regression analysis manually on large datasets with multiple independent variables can be tedious. An automated system for regression analysis will be of great help for researchers as well as non-expert users. Thus, the objective of this research is to design and develop an automated curve fitting system. As outcome, a curve fitting system named "CurFi" was developed that uses linear regression models to fit a curve to a dataset and to find out the best fit model. The system facilitates to upload a dataset, split the dataset into training set and test set, select relevant features and label from the dataset; and the system will return the best fit linear regression model after training is completed. The developed tool would be a great resource for the users having limited technical knowledge who will also be able to find the best fit regression model for a dataset using the developed "CurFi" system.

**KEYWORDS**
best fit, curve fitting, least squares, machine learning, regression, regression analysis

**JEL CLASSIFICATION**
Applied science for engineering

## 1 | INTRODUCTION

Curve fitting is a preliminary activity to many techniques used to model and solve production problems such as simulation, predictive modeling, and statistical inference.[1] Scientists and engineers often want to represent empirical data using a model based on mathematical equations. With the correct model and calculus, one can determine important characteristics of the data, such as the rate of change anywhere on the curve (first derivative), the local minimum and maximum points of the function (zeros of the first derivative), and the area under the curve (integral). The goal of data (or curve) fitting is to find the parameter values that most closely match the data. To find the values of the model's parameters that yield the curve closest to the data points, one must define a function that measures the closeness between the data and the model. This function depends on the method used to do the fitting, and the function is typically chosen as the sum of the squares of the vertical deviations from each data point to the curve (the deviations are first squared and then added up to avoid cancellations between positive and negative values).[2] This technique is known as the least square method and it is the oldest curve fitting technique which later in the 19th century came to be known as regression analysis.[3] The assumption in the least square method is that the model has normally distributed errors.[4] When the errors in the system under investigation are not normally distributed, however, the coefficients generated by the least squares approach are not necessarily the most likely.[5] To address this problem, robust regression methods have been developed. Among all the approaches to robust regression, the least absolute deviations (LADs) method, or L1-norm, is considered conceptually the simplest one.[4] In the least absolute deviations (LADs) method, the absolute values of the vertical deviations from each data point to the curve are taken. These are known as absolute deviations and the best fit curve is defined here to be that which minimizes the maximum absolute deviation between the fitted curve and the given data.[6] In either method, the best fit model is defined as the model that most closely matches the data.

Regression analysis is a statistical technique to determine the correlations between two or more variables having cause-effect relations and to make predictions for the topic using the relation.[7] The analysis yields a predicted value for the criterion or dependent variable resulting from a linear combination of predictors or independent variables.[8] Linear regression is the best known and most easily understood form of regression analysis.[9] Currently, regression models are being applied in many disciplines like linguistics, sociology, history, astronomy, and geography.[10-12]

The task of deciding which of several regression models describe a dataset best is a common problem in many disciplines. In many cases, normal statistical calculations are used to find a best fit model or a program is written based on these statistical calculations. But this approach is often time consuming and tedious. Moreover, various math and statistical analysis softwares such as Mathcad, Excel Data Analysis package and SPSS (SPSS is short for Statistical Product and Service Solutions and it is a statistical software suite used for complex statistical data analysis) are currently available. These tools are also extensively used to conduct regression analysis. But most of these tools have specific rules and specifications that the user must follow to conduct regression analysis. So the user may require substantial knowledge of regression analysis before using these tools.

Recently automated machine learning tools have gained popularity and various web applications have been created where users can upload a dataset, tweak the parameters of a particular model and then train the machine learning model. This type of web application has substantially reduced human effort to create and run machine learning models.[13] Linear regression algorithm is the simplest machine learning algorithm and it can be integrated into a web app.[14] Deploying a machine learning algorithm as a web service hosted on the local system or in the cloud is a convenient way to automate the process of training the model and making prediction from the dataset.[15] Therefore, the objective of this research is to develop an automated system to find patterns in the dataset automatically, get insights and make future predictions from the dataset. To attain this objective, a web application system was developed so that users can upload their dataset and the system returns the best-fit regression model along with the accuracy scores. Also, a library named "curfi.js" was published alongside the web application, so that any developer can use the library to create more robust automated curve fitting web applications in the future.[16]

## 2 | LITERATURE REVIEWS

Some previous works will be discussed to understand the premise of this work. Fumo et al.[17] have described how to model the residential energy consumption using linear regression. In this study, Fumo et al. have conducted a simple linear regression, a multiple linear regression as well as a quadratic linear regression analysis to model the energy consumption of a house with daily and hourly data. The researchers used "Mathcad" and "Excel Data Analysis package" to conduct the regression analysis. But to build the regression models and compare the accuracy among them, the regression functions provided by these tools had to be manually written and the required parameters needed to be passed to those functions to get the desired results. There are various mathematical functions provided by these math softwares. But these mathematical functions must be written and run manually to get the desired result. Olaniyi et al.[18] have demonstrated how to discover knowledge from databases. Organizations collect massive amounts of data and store it in data warehouses. But most of this data remains unused and organizations do not often know how to extract useful information from massive amounts of data. So the researchers in this study applied regression analysis for stock price prediction to demonstrate that regression analysis is suitable for knowledge discovery from data. A data mining software tool was developed which employed the use of regression analysis through the use of time series data to predict future stock market prices. This type of data mining tool usually requires a significant amount of time and effort to build and maintain. Bradshaw et al.[19]

have developed a multiple linear regression model to predict VO2max based on non-exercise (N-EX) data. VO2max is a measure of the maximum oxygen uptake of a person before or after intense exercise. VO2max is one of the parameters to determine a person's health condition. A multiple linear regression equation is generated considering total 5 parameters as the independent variables and VO2max as the dependent variable. The multiple linear regression model was derived by statistical technique that is normally used to get the best fit model for a given dataset. These statistical techniques require sophisticated mathematical calculations which take a considerable amount of time. Heim et al.[20] used multiple linear regression analysis to predict the level of neuroendocrine stress response in adult women who have experienced childhood abuse in the past or major life events recently. Studies have shown that childhood trauma and major life events are correlated with the hormonal system of central nervous system of humans. This system is also known as the neuroendocrine stress response system. So in this study, the researchers decided to measure the neuroendocrine stress response in adult women with respect to some predictors like demographic variables, childhood trauma, adulthood trauma, major life events in the past year and daily hassles in the past month and so forth. Then the researchers used this data to develop a multiple linear regression model that can reveal which predictor is the most influential factor in neuroendocrine stress response. For this SPSS was used. This software tool provides functionalities for conducting linear regression analysis but the output generated by this tool is not easily interpretable to a non-expert user. In other words, for using this statistical analysis software, one has to be well acquainted with the process of regression analysis and the mathematical parameters associated with regression analysis. Chandrashekar Murthy et al.[21] used a multiple linear regression model to predict the water demand for domestic purpose of a house. A progressive web app was developed to take the user data and show the prediction of water usage as output. The multiple linear regression model was implemented using python and it was integrated with a web app to generate the output. Implementing multiple linear regression model using programming language like python requires substantial programming skill and embedding the model into a web app can automate the process. Barhmi et al.[22] developed multiple linear regression and artificial neural network models to forecast wind speed in north and south regions of Morocco for 3 years. The equations for the models were derived manually without the help of any automated tool. Patel et al.[23] used both linear and nonlinear regression models to predict the surface roughness level of a machined surface and to compare it to the measured values. This work would become easier for the researchers if an automated tool was used to perform the regression analysis. Nyarko-Boateng et al.[24] presented a machine learning approach to predict the actual location of a fiber cable fault in an underground optical transmission link. The researchers used single-layer perceptron neural network and a simple linear regression model to predict the actual location of the fault in the underground optical network with high accuracy. Python sci-kit learn library was used to build the simple linear regression model. But before using sci-kit learn library, one must have enough knowledge of python programming and machine learning models.

Linear regression models are used intensively in multiple disciplines including engineering science, medical science, stock market prediction, energy sector and so forth. After summarizing these works, a few concerns have come out from the literature survey. Researchers have used various software packages such as Mathcad, Microsoft Excel and SPSS to conduct linear regression analysis. These software tools include mathematical functions which have to be written and run to conduct regression analysis. Moreover, understanding the output generated by these tools is not always feasible for non-expert users. Researchers have also applied normal statistical techniques to derive the best fit linear regression model manually which is often time consuming and tedious. Also programming languages like Python and frameworks like sci-kit learn have been used to build linear regression models which require enough programming skills. So an automated tool for linear regression analysis will be of great help in these scenarios where time and effort could be saved and it will be also useful for non-expert users to easily interpret the result of the analysis.

## 3 | METHODOLOGY

This section first describes how to find the best fit model. Then it focuses on how to automate the process along with the help of some algorithms. After that, the chapter discusses the design and development of the system. Lastly, the chapter focuses on the development of the library which was published in npm.

### 3.1 | Finding the best fit model

The most widely used linear regression models are described in Sections 3.1.1–3.1.6. Then a generalized equation is formed from those linear regression equations on the Section 3.1.7.

### 3.1.1 | Simple linear regression

The simple linear regression model is a straight line fit to a dataset. In simple linear regression, there is only one dependent variable($y$) and one independent variable($x$). The mathematical expression of this model is shown in Equation (1).

$$y = a_0 + a_1 x + error, \tag{1}$$

where $a_0$ and $a_1$ are the coefficients representing the intercept and slope respectively. Here *error* is the residual error which is a measure of deviation of the model from the dataset.

### 3.1.2 | Multiple linear regression

A simple linear regression model represents the relationship between one feature and one dependent variable. But multiple linear regression is a mathematical technique used to model the relationship between multiple independent predictor variables and a single dependent outcome variable.[25] The equation for the multiple linear regression model can be described as Equation (2).

$$y = a_0 + a_1 x_1 + a_2 x_2 + \cdots + a_m x_m + error. \tag{2}$$

Here, the parameters are denoted by $a_0, a_1, a_2, \ldots, a_m$ and the residual error is denoted by *error*. The goal is to find out the best values for these parameters to find the "best fit" model for a given dataset.

### 3.1.3 | Polynomial linear regression

The polynomial linear regression model uses a higher-order polynomial equation where the independent variable $x$ has higher-order power. For example, an $m$th order polynomial linear regression model is denoted by the following Equation (3).

$$y = a_0 + a_1 x + a_2 x^2 + \cdots + a_m x^m + error. \tag{3}$$

The parameters of this model are $a_0, a_1, a_2, \ldots, a_m$ and the residual error is denoted by *error*. The goal is to find the best values of these parameters which will give us the "best fit" model for a given dataset.

### 3.1.4 | Logarithmic linear regression

This regression model can incorporate multiple independent variables and one dependent variable. But the only difference is that the independent variables are incorporated into this model as the natural logarithm of $x$ where $x$ is an independent variable. The equation of this model is shown in Equation (4).

$$y = a_0 + a_1 \ln x_1 + a_2 \ln x_2 + \cdots + a_m \ln x_m + error, \tag{4}$$

where the parameters are $a_0, a_1, a_2, \ldots, a_m$ and the residual error is represented by *error*.

### 3.1.5 | Exponential linear regression

The exponential linear regression model can incorporate multiple independent variables and one dependent variable. But the independent variables are used in the model as the power of the exponent $e$. The equation of this model is shown in Equation (5).

$$y = a_0 + a_1 e^{x_1} + a_2 e^{x_2} + \cdots + a_m e^{x_m} + error, \tag{5}$$

where $a_0, a_1, a_2, \ldots, a_m$ are parameters of the model and *error* is the residual error.

### 3.1.6 | Sinusoidal linear regression

The sine and cosine functions are periodic functions and they can be used to construct linear regression models which are called "Sinusoidal linear regression" models. To construct these models, both sine and cosine functions can be used; there is no clear-cut convention for choosing either function and the results will be identical for both functions. The linear regression model equation for sine function is expressed as Equation (6).

$$y = A_0 + C_1 \sin(x + \theta). \tag{6}$$

In Equation (6), $x$ is the independent variable and $y$ is the dependent variable. The parameters required to build the sinusoidal model are $A_0$, $C_1$, and $\theta$. $A_0$ is the mean value that denotes the average height of the sinusoidal function above the x-axis. $C_1$ is the amplitude of the sine wave which denotes the height of oscillation. Finally, $\theta$ is called the phase shift which denotes the extent of the shift of the wave horizontally.

But Equation (6) has non-linear characteristics because of the $\theta$ term. So this equation has to be converted in the form of a linear regression model like the previously described models. To achieve the linear regression form, the $C_1 \sin(x + \theta)$ part of Equation (6) will be converted as follows

$$\begin{aligned} C_1 \sin(x + \theta) &= C_1(\sin x \cos \theta + \cos x \sin \theta) \\ &= A_1 \sin x + A_2 \cos x. \end{aligned} \tag{7}$$

So the final equation in the linear regression model form will look like Equation (8)

$$y = A_0 + A_1 \sin x + A_2 \cos x. \tag{8}$$

So now Equation (8) has 3 parameters $A_0$, $A_1$, and $A_2$. But the initial non-linear Equation (6) had 3 parameters $A_0$, $C_1$, and $\theta$. $C_1$ and $\theta$ can be found using $A_1$ and $A_2$

$$C_1 = \sqrt{A_1^2 + A_2^2}, \tag{9}$$

$$\theta = \tan^{-1} \frac{A_2}{A_1}. \tag{10}$$

So after linearizing the non-linear Equation (6), the Equation (8) got 3 parameters $A_0$, $A_1$, and $A_2$. After determining the best values of the 3 parameters $A_0$, $A_1$, and $A_2$, the Equation (6) parameters $C_1$ and $\theta$ can be determined from $A_1$ and $A_2$ using Equations (9) and (10).

### 3.1.7 | Finding the generalized equation

The procedure for finding the best values for the parameters is quite the same for all linear regression models. At first, the previously described linear regression models can be expressed by the following general Equation (11).

$$y = a_0 z_0 + a_1 z_1 + a_2 z_2 + \cdots + a_m z_m + error. \tag{11}$$

Here, $z_0, z_1, z_2, \ldots, z_m$ are called basis functions. For each linear regression model, the basis functions will be different. For instance, the basis functions for multiple linear regression models will be $z_0 = 1, z_1 = x_1, z_2 = x_2, \ldots, z_m = x_m$. Similarly, the basis functions for a polynomial linear regression model will be $z_0 = 1, z_1 = x, z_2 = x^2, \ldots, z_m = x^m$.

At first, the equation of the sum of squared residuals can be expressed as Equation (12). In Equation (12) $i$ denotes individual data point and we are calculating residual error for $n$ data points.

$$S_r = \sum_{i=1}^{n}(y_i - a_0 z_{0i} - a_1 z_{1i} - a_2 z_{2i} - \cdots - a_m z_{mi})^2. \tag{12}$$

Then the derivative of $S_r$ with respect to each of the parameters $a_0, a_1, a_2, \ldots, a_m$ needs to be derived as shown in Equations (13) to (16).

$$\frac{\partial S_r}{\partial a_0} = -2 \sum z_{0i}(y_i - a_0 z_{0i} - a_1 z_{1i} - a_2 z_{2i} - \cdots - a_m z_{mi}), \tag{13}$$

$$\frac{\partial S_r}{\partial a_1} = -2 \sum z_{1i}(y_i - a_0 z_{0i} - a_1 z_{1i} - a_2 z_{2i} - \cdots - a_m z_{mi}), \tag{14}$$

$$\frac{\partial S_r}{\partial a_2} = -2 \sum z_{2i}(y_i - a_0 z_{0i} - a_1 z_{1i} - a_2 z_{2i} - \cdots - a_m z_{mi}), \tag{15}$$

$$\vdots$$

$$\frac{\partial S_r}{\partial a_m} = -2 \sum z_{mi}(y_i - a_0 z_{0i} - a_1 z_{1i} - a_2 z_{2i} - \cdots - a_m z_{mi}). \tag{16}$$

After finding out the derivatives, Equations (13) to (16) will be set equal to 0. So total $m$ linear equations will be created. This system of linear equations can be solved to find out the best values of the parameters. These values will give us the linear regression model which will "best fit" the data.

$$\sum (z_{0i})^2 a_0 + \sum (z_{0i})(z_{1i}) a_1 + \sum (z_{0i})(z_{2i}) a_2 + \cdots + \sum (z_{0i})(z_{mi}) a_m = \sum z_{0i} y_i, \tag{17}$$

$$\sum (z_{0i})(z_{1i}) a_0 + \sum (z_{1i})^2 a_1 + \sum (z_{1i})(z_{2i}) a_2 + \cdots + \sum (z_{1i})(z_{mi}) a_m = \sum z_{1i} y_i, \tag{18}$$

$$\sum (z_{0i})(z_{2i}) a_0 + \sum (z_{1i})(z_{2i}) a_1 + \sum (z_{2i})^2 a_2 + \cdots + \sum (z_{2i})(z_{mi}) a_m = \sum z_{2i} y_i, \tag{19}$$

$$\vdots$$

$$\sum (z_{0i})(z_{mi}) a_0 + (z_{1i})(z_{mi}) a_1 + \sum (z_{2i})(z_{mi}) a_2 + \cdots + \sum (z_{mi})^2 a_m = \sum z_{mi} y_i. \tag{20}$$

This system of linear equations comprising of Equations (17) to (20) can also be expressed in matrix notation like Equation (21).

$$\begin{bmatrix} \sum z_{0i}^2 & \sum z_{0i} z_{1i} & \sum z_{0i} z_{2i} & \cdots & \sum z_{0i} z_{mi} \\ \sum z_{0i} z_{1i} & \sum z_{1i}^2 & \sum z_{1i} z_{2i} & \cdots & \sum z_{1i} z_{mi} \\ \sum z_{0i} z_{2i} & \sum z_{1i} z_{2i} & \sum z_{2i}^2 & \cdots & \sum z_{2i} z_{mi} \\ \vdots & & & \ddots & \\ \sum z_{0i} z_{mi} & \sum z_{1i} z_{mi} & \sum z_{2i} z_{mi} & \cdots & \sum z_{mi}^2 \end{bmatrix} \begin{bmatrix} a_0 \\ a_1 \\ a_2 \\ \vdots \\ a_m \end{bmatrix} = \begin{bmatrix} \sum z_{0i} y_i \\ \sum z_{1i} y_i \\ \sum z_{2i} y_i \\ \vdots \\ \sum z_{mi} y_i \end{bmatrix}. \tag{21}$$

If we consider $A = \begin{bmatrix} a_0 \\ a_1 \\ a_2 \\ \vdots \\ a_m \end{bmatrix}$ and $Z = \begin{bmatrix} \sum z_{0i}^2 & \sum z_{0i} z_{1i} & \sum z_{0i} z_{2i} & \cdots & \sum z_{0i} z_{mi} \\ \sum z_{0i} z_{1i} & \sum z_{1i}^2 & \sum z_{1i} z_{2i} & \cdots & \sum z_{1i} z_{mi} \\ \sum z_{0i} z_{2i} & \sum z_{1i} z_{2i} & \sum z_{2i}^2 & \cdots & \sum z_{2i} z_{mi} \\ \vdots & & & \ddots & \\ \sum z_{0i} z_{mi} & \sum z_{1i} z_{mi} & \sum z_{2i} z_{mi} & \cdots & \sum z_{mi}^2 \end{bmatrix}$ and $Y = \begin{bmatrix} \sum z_{0i} y_i \\ \sum z_{1i} y_i \\ \sum z_{2i} y_i \\ \vdots \\ \sum z_{mi} y_i \end{bmatrix}$. Then Equation (21) becomes like Equation (22).

$$Z \cdot A = Y. \tag{22}$$

If $Z$ is brought to the right side, essentially deriving its inverse matrix, and multiplying this inverse matrix with the vector $Y$, then the vector $A$ can be found in this way as shown in Equation (23).

$$A = Z^{-1} \cdot Y. \tag{23}$$

The purpose of Algorithm 1 is to create a linear regression model based on the dataset's features and label and also the *type* parameter which refers to one of the six types of linear regression models. The arguments are *trainX*, *trainY*, *type* and *highestOrder* where *trainX* is feature data points and *trainY* is label data points of the dataset, *type* denotes the type of linear regression model that needs to be created and *highestOrder* denotes the order of the polynomial regression model. The algorithm returns the coefficient vector A or *error* if coefficient vector A can not be created. Lines 2–8 is the initialization part where matrices *x*, *y*, and *a* are initialized. The matrices *x* and *y* take values from *trainX* and *trainY* respectively and store them in a proper format. The matrix *a* stores the calculated values of Z and Y matrices which will later be used to generate the coefficient vector A. Line 9–24 generate the values of the Z matrix and store these values in the matrix *a*. Line 25–34 generate the values of the Y matrix and store them in the matrix *a*. The Z and Y matrices are required to create the coefficient vector A that will give us the best fit linear regression model. Line 35–38 generates the coefficient vector A from the inverse matrix of Z and Y. Finally line 39 returns the coefficient vector A.

## 3.2 | Model selection by computing R squared value

To automatically select the best model we first need to define a way to compare the models. The best way to compare several regression models is to calculate the accuracy of these models and then sort them to get the best fit model for a given dataset. For the accuracy we can compute the R squared ($r^2$) value for a regression model for a given dataset. To do this, we return to the original dataset and determine the total sum of the squares around the mean for the dependent variable (in our case, *y*). This quantity is designated $S_t$. This is the magnitude of the residual error associated with the dependent variable prior to regression.[26] After performing the regression, we can compute $S_r$, the sum of the squares of the residuals around the regression line. This characterizes the residual error that remains after the regression.[26] The difference between the two quantities, $S_t - S_r$, quantifies the improvement or error reduction due to conducting regression analysis rather than as an average value. Because the magnitude of this quantity is scale-dependent, the difference is normalized to $S_t$ to provide Equation (24).

$$r^2 = \frac{S_t - S_r}{S_t}, \qquad (24)$$

where $r^2$ is called the coefficient of determination and *r* is the correlation coefficient ($\sqrt{r^2}$). For a perfect fit, $S_r = 0$ and $r = r^2 = 1$, signifying that the model explains 100% of the variability of the data. For $r = r^2 = 0$, $S_r = S_t$ and the fit represents no improvement after conducting regression analysis.

With the $r^2$ value, we can compare the regression models for a given dataset. When $r^2$ is close to 1 (or equals to 1) the model is good and represents significant improvement after regression analysis. We run the given dataset for all the regression models discussed in previous subsections and compute $r^2$ values for all these models for that dataset. After sorting these models based on their $r^2$ values, the regression model with the greater $r^2$ value is the best fit model. This best fit model is the model that best describes the relationship of the given dataset.

This automated process is run in the web app using the *AutoTrain* function that takes *trainX*, *trainY*, *testX*, *testY*, *highestOrder* as function parameters. Here, *trainX* is the training dataset's features list, namely the list of independent variable($x_i$) values. *trainY* is the training dataset's label list, namely the list of dependent variable (*y*) values. Similarly *testX* and *testY* are the test dataset's features list and label list respectively. Here, the "training dataset" and "test dataset" simply denotes a feature that is implemented in the web application. In this case, the linear regression model has been considered as the simplest machine learning model and the user can split the dataset into training and test dataset. The *highestOrder* is the value for *m* which is the order and used by the polynomial regression model. The *AutoTrain* function generates each of the linear regression models using Algorithm 1, discussed on Section 3.1.7, that takes *trainX*, *trainY*, *type* and *highestOrder* as parameters and returns a trained linear regression model along with its coefficient vector A for both training and test datasets. The coefficient vector A is used to calculate the $r^2$ score of each linear regression model which indicates how well the model fits the dataset. *AutoTrain* function then sorts these models based on their $r^2$ scores and returns the best fit model. This best fit model along with other models are then shown in the UI in sorted order to the user.

The equations for these regression models along with the values of $a_0, a_1, a_2, \ldots, a_m$ are also shown in the UI so that users can easily get the insight and relationship from the dataset.

**Algorithm 1.** Create a linear regression model based on the *type* parameter

**Input:** *trainX*, the feature data points
*trainY*, the label data points of the dataset and
*type*, the regression type
*highestOrder*, order of polynomial regression model
**Output:** the coefficient vector *A*

1: **procedure** CREATE_MODEL(*trainX*, *trainY*, *type*, *highestOrder*)
2:    $n \leftarrow$ no of data points
3:    $x0 \leftarrow$ an array of length *n* filled with 1
4:    $x \leftarrow$ feature values from *trainX* in a proper format
5:    $y \leftarrow$ feature values from *trainY* in a proper format
6:    insert *x0* at the beginning of *x*
7:    *order* $\leftarrow$ order of the linear regression model to be constructed
8:    $a \leftarrow$ an array of length *order* + 2 filled with 0
9:    **for** $i \leftarrow 1$ to *order* + 1 **do**
10:       **for** $j \leftarrow 1$ to $i$ **do**
11:          **if** *type* == "polynomial" **then**
12:             $k \leftarrow i + j - 2$
13:          **end if**
14:          $sum \leftarrow 0$
15:          **for** $l \leftarrow 0$ to $n - 1$ **do**
16:             **if** *type* == "polynomial" **then**
17:                $sum \leftarrow sum + (x[1][l])^k$
18:             **else**
19:                $sum \leftarrow sum + x[i-1][l] * x[j-1][l]$
20:             **end if**
21:          **end for**
22:          $a[i][j] \leftarrow sum$
23:          $a[j][i] \leftarrow sum$
24:       **end for**
25:       $sum \leftarrow 0$
26:       **for** $l \leftarrow 0$ to $n - 1$ **do**
27:          **if** *type* == "polynomial" **then**
28:             $sum \leftarrow sum + y[0][l] * (x[1][l])^{i-1}$
29:          **else**
30:             $sum \leftarrow sum + y[0][l] * x[i-1][l]$
31:          **end if**
32:       **end for**
33:       $a[i][order + 2] \leftarrow sum$
34:    **end for**
35:    $Z \leftarrow$ extract values of matrix *Z* from matrix *a*
36:    $Y \leftarrow$ extract values of matrix *Y* from matrix *a*
37:    $Zinv \leftarrow$ calculate inverse matrix of matrix *Z*
38:    $A \leftarrow$ calculate *A* from *Zinv* and *Y*
39:    **return** *A*
40: **end procedure**

## 3.3 | Design and development of the system

A web-based system (*CurFi*) was developed so that users can use the automated regression tool without installing any software or program. Any user can access the web application by using the address of the website from a browser.[27]

The system will help the researchers with little technical knowledge be able to easily find the best model that fits a dataset. The web application system contains all the features needed for a user to navigate the system, select required features and target labels from the dataset and find out which model fits the data best, the accuracy of that best-fit model, and also what other models might fit the dataset.

The system takes in a dataset as input from the user in CSV (comma separated values) format. The user then also selects the features and target label associated with that dataset for fitting the model. The system then converts that CSV file and turns it into a two-dimensional array-like data structure for calculation on the data points more conveniently. This converted dataset is then split into two datasets (1) training dataset, (2) test dataset. The train-test split percentage can be set by the user from the user interface. The training dataset is then fit to the six different linear regression models.

The UI is designed in such a way that it takes very little effort for any user to use the system. After going to the website, the user can upload a dataset using the "Choose File" button. The user can also select the train-test split percentage in the UI. After that, the user needs to choose the features and label from the UI for the automated regression analysis. Then, the user can click the "train" button and then the system will automatically calculate and fit all the regression models. After fitting all the regression models and then computing and comparing these models the models along with their $r^2$ score and equations for the models are shown on the UI so that the user can better understand the relationships in the dataset. The models are shown on the UI in sorted order according to their $r^2$ scores. This means that the model that is shown in top is the best fit model. Some snapshots of UI are shown in Figures 1–6.

The UI is designed using the HTML markup language. And the site is designed using the CSS language. For interactions within the website like button clicks, dataset uploading and so forth. JavaScript is used. The website is then deployed to a cdn server that sends the site to any user who requests for the website.

## 3.4 | Publishing the library

Apart from developing a web application system we also published a library named "curfi.js."[16] This library is written completely in JavaScript language. This is also the same library that was used for developing the web application system. We published the library on npm which is a JavaScript library distribution platform so that anyone can use our library to build more robust and automated systems for the web.

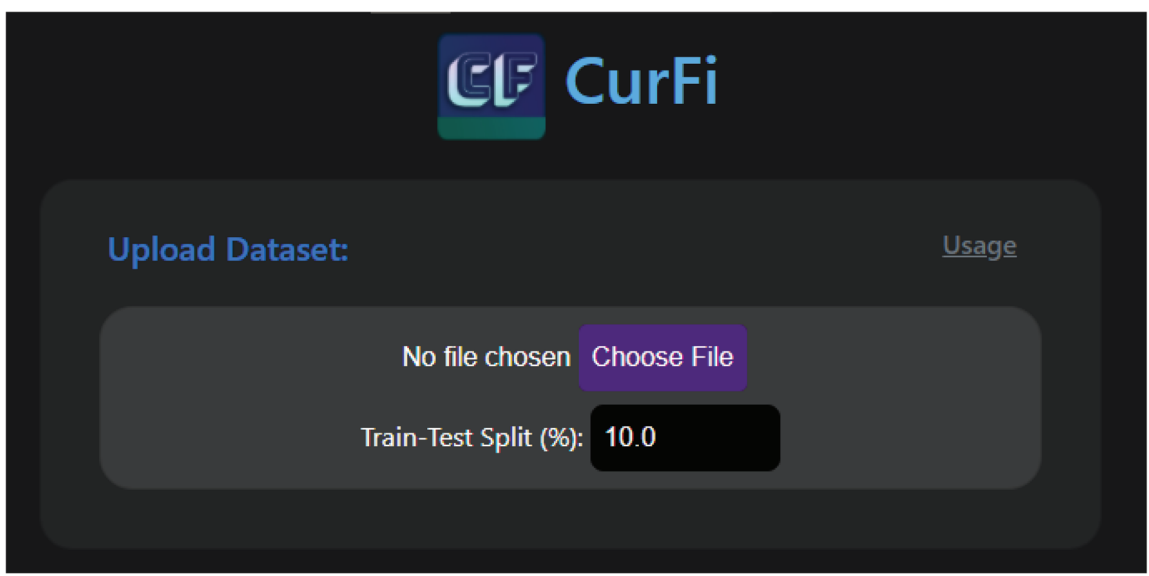

**FIGURE 1** UI for uploading the dataset

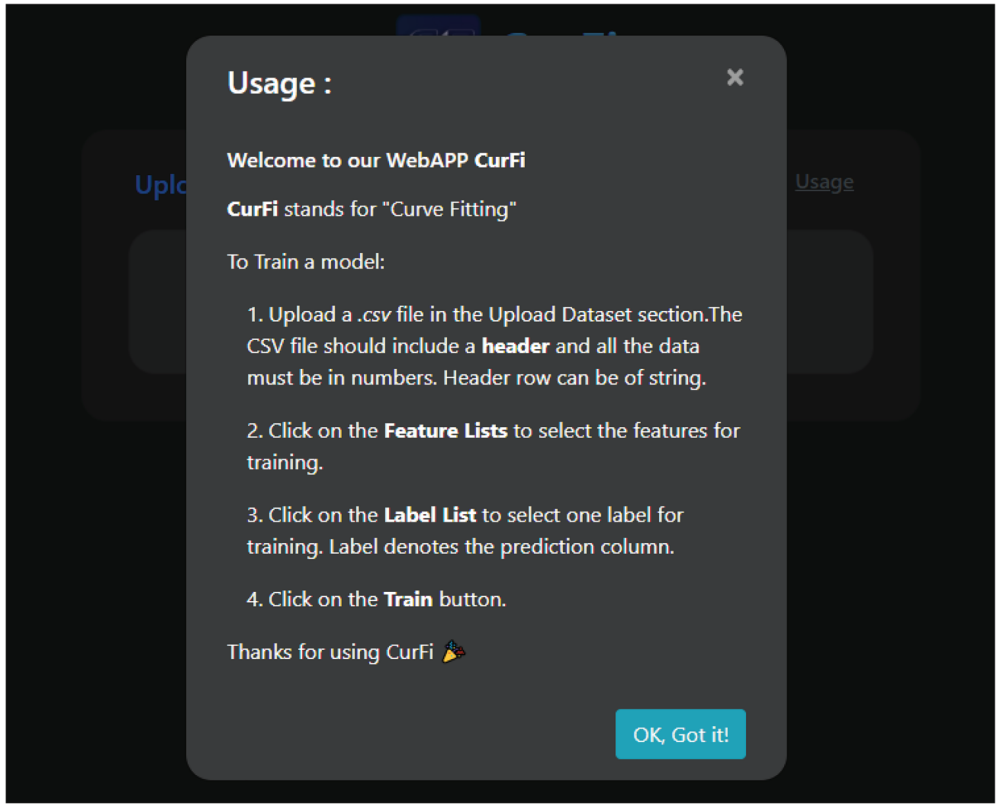

**FIGURE 2**   UI for describing the usage

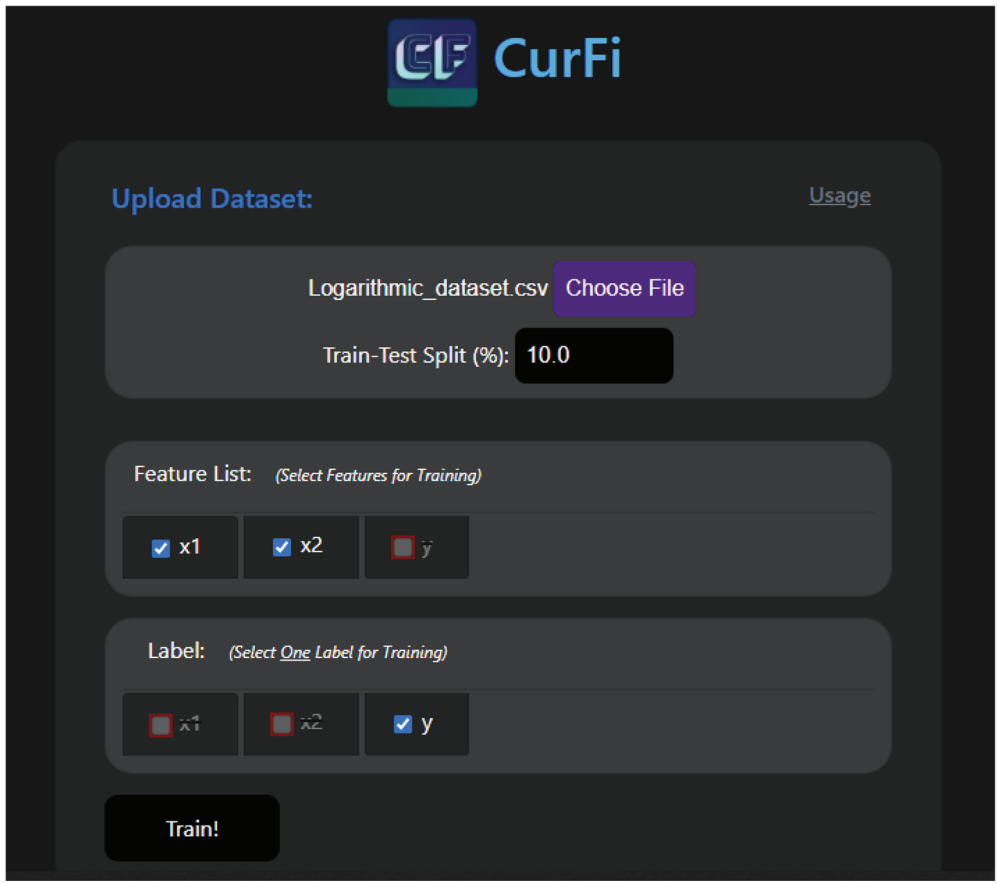

**FIGURE 3**   UI for selecting feature list and label

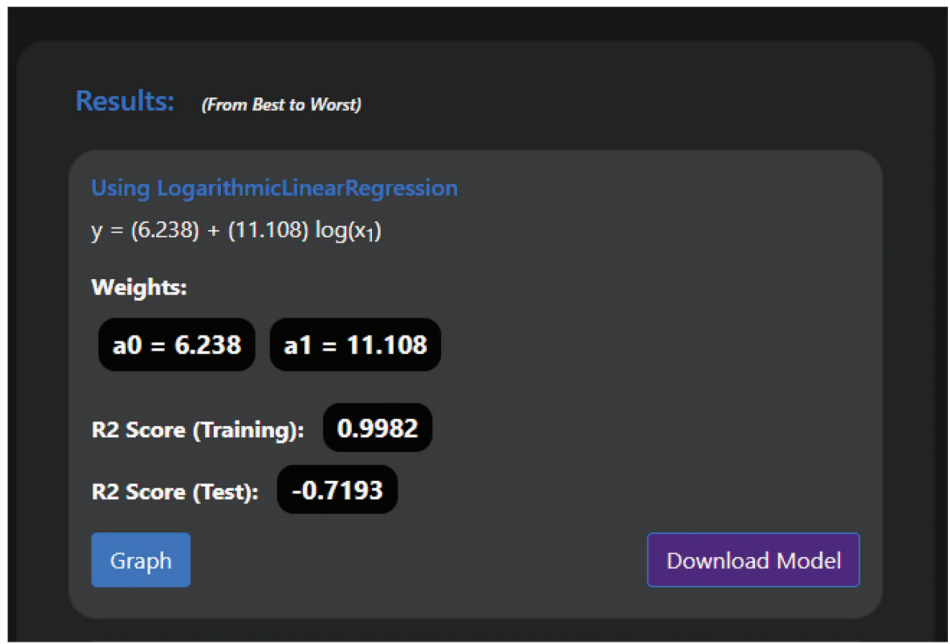

**FIGURE 4** UI for showing the equation, parameters, and $r^2$ score of a regression model.

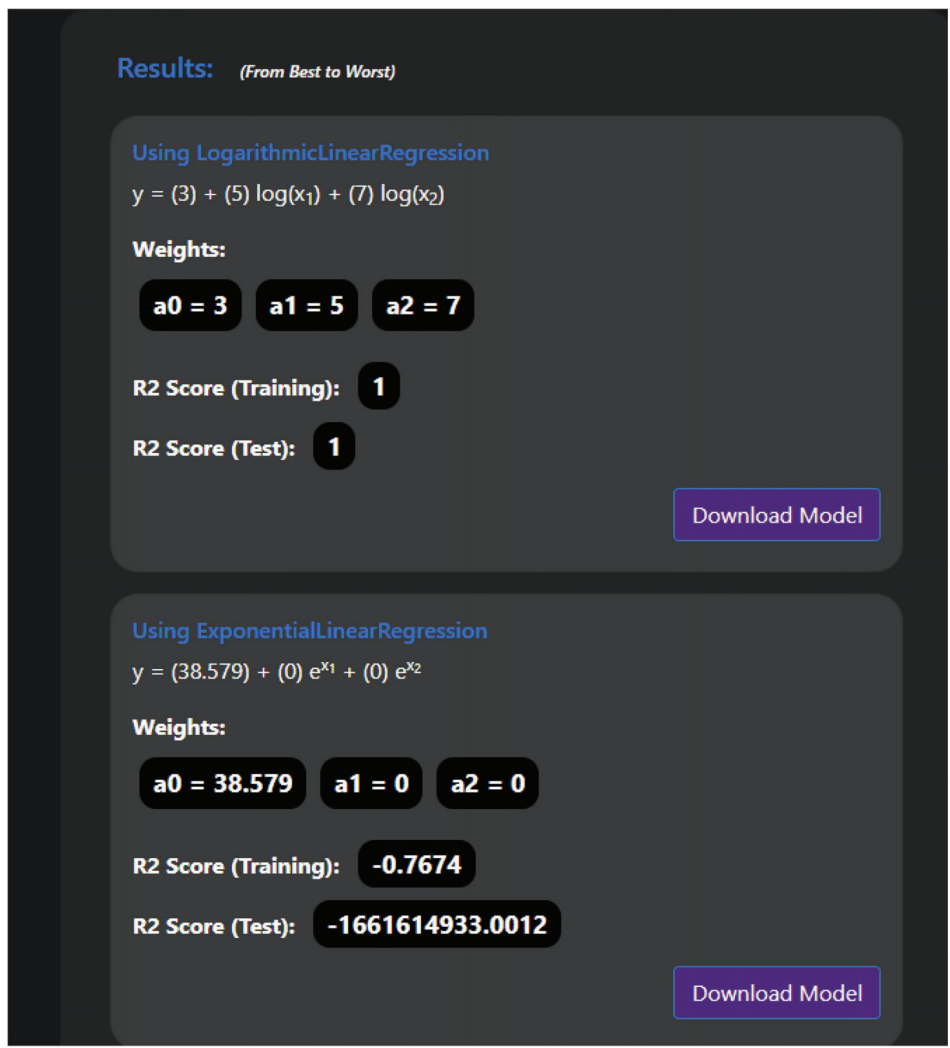

**FIGURE 5** UI for describing the regression models sorted from best to worst.

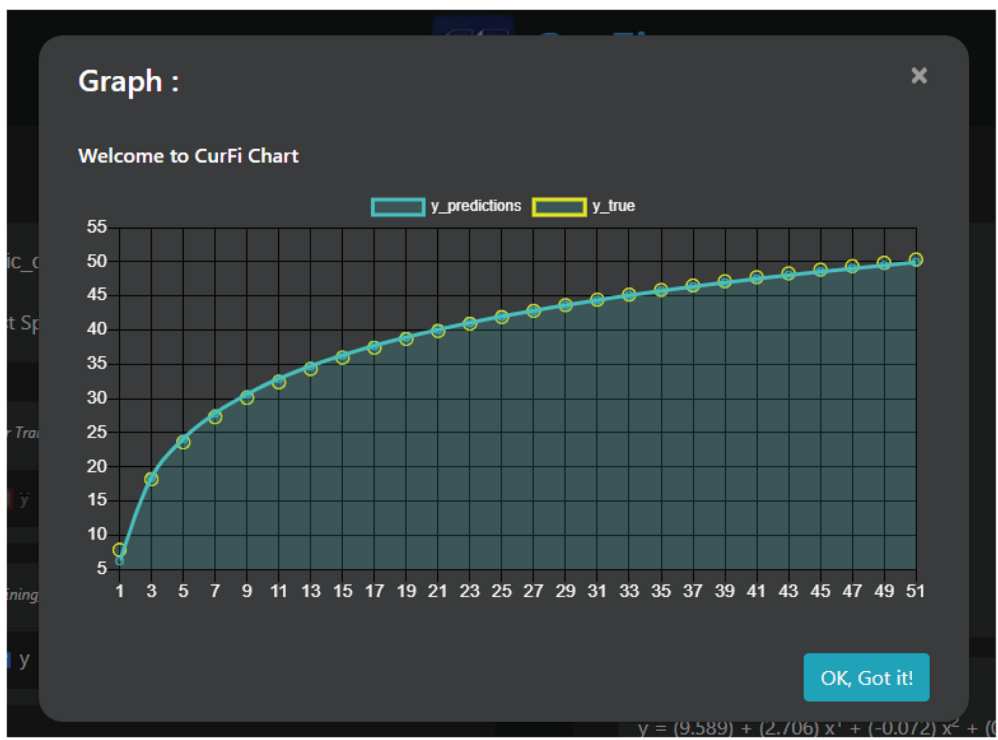

**FIGURE 6** UI for showing the graphical visualization of a regression model

## 4 | EVALUATING THE SYSTEM

The developed system is evaluated based on three objectives. The first one is to evaluate the effectiveness of the system to find out an accurate relationship between one or more independent variables and a dependent variable. Then the ability to find the "best fit" model automatically is evaluated. After that the ability of the system to find patterns in a real-world dataset is also evaluated.

### 4.1 | Exploring relationship between independent and dependent variables

For performing the evaluation of the system, some datasets were created according to the linear regression models that are implemented in the system. The datasets contained only numerical variables. These datasets were then uploaded to the system and trained to find out if the system is able to find the relationship between the independent and dependent variables accurately. At least one dataset was created for each of the linear regression models present in the system.

In all cases, the system reproduced the linear regression equations with a very high $r^2$ score, almost close to 1. Table 1 shows the performance of the system for these prepared datasets.

**TABLE 1** Performance analysis for the system

| Regression type | Proposed relationship (equation) | Equation by system | Training R2 score | Test R2 score |
| --- | --- | --- | --- | --- |
| Linear regression | $2 + 3x$ | $2 + 3x$ | 1.00 | 1.00 |
| Multiple linear regression | $15 + 9x_1 - 6x_2$ | $14.805 + 8.874x_1 - 5.842x_2$ | 0.99 | 0.99 |
| Exponential regression | $2 + 3e^{x_1} + 8e^{x_2}$ | $2 + 3e^{x_1} + 8e^{x_2}$ | 1.00 | 1.00 |
| Polynomial regression | $3 + 4x + 8x^2$ | $3 + 4x + 8x^2$ | 1.00 | 1.00 |
| Logarithmic regression | $-1.57 + 4.4 \ln(x_1) + 3.6 \ln(x_2)$ | $-1.569 + 4.36 \ln(x_1) + 3.59 \ln(x_2)$ | 0.98 | 0.96 |
| Sinusoidal regression | $3 + 4 \sin(x + 5)$ | $3 + 4 \sin(x + 5)$ | 1.00 | 1.00 |

As shown in Table 1 the system is able to produce the relationship between independent and dependent variables with $r^2$ score close to 1. This means that the system can accurately find out the relationship between one or more independent variables and the dependent variable.

## 4.2 | Finding the best fit model

To evaluate the system's ability to find the best fit model, some demo datasets were created based on each of the linear regression models. Also an open source dataset called "Breast Cancer Wisconsin" was used to evaluate the system.[28] All variables in these datasets were of numerical type. These datasets were uploaded and trained to observe if the system is able to return the best fit model.

For each of the datasets, the system was able to return the linear regression models in sorted order according to their $r^2$ scores. The best fit model is the one with the highest $r^2$ score. Since the linear regression models were returned in sorted order, the best fit model was shown on top.

For each of the demo datasets as well as the open source dataset, the system was able to return the best fit model accurately. This implies that the system is able to find the best fit model for any numerical dataset which has a linear relationship between the independent variables and the dependent variable.

**FIGURE 7** Selecting the feature list and label of the dataset

## 4.3 | Finding pattern in open source datasets

Besides demo datasets an open source dataset named "Breast Cancer Wisconsin"[28] was used to evaluate the system. The dataset has the independent variables "Sample code number," "Clump Thickness," "Uniformity of Cell Size," "Uniformity of Cell Shape," "Marginal Adhesion," "Single Epithelial Cell Size," "Bare Nuclei," "Bland Chromatin," "Normal Nucleoli," and "Mitoses." Among these independent variables, "Sample code number" was excluded because it is just an ID number that will not contribute to the regression analysis. Each independent variable except "Sample code number" has the domain 1–10. The dependent variable is "class" which is a binary variable representing breast cancer type. It has values 2 and 4 where 2 indicates "benign" and 4 indicates "malignant."

After this dataset was uploaded and trained, the system was able to find the accurate value of the dependent variable with a very high $r^2$ score. For predicting the cancer type, the logarithmic linear regression model was returned as the best fit model. The results are shown in Figures 7 and 8.

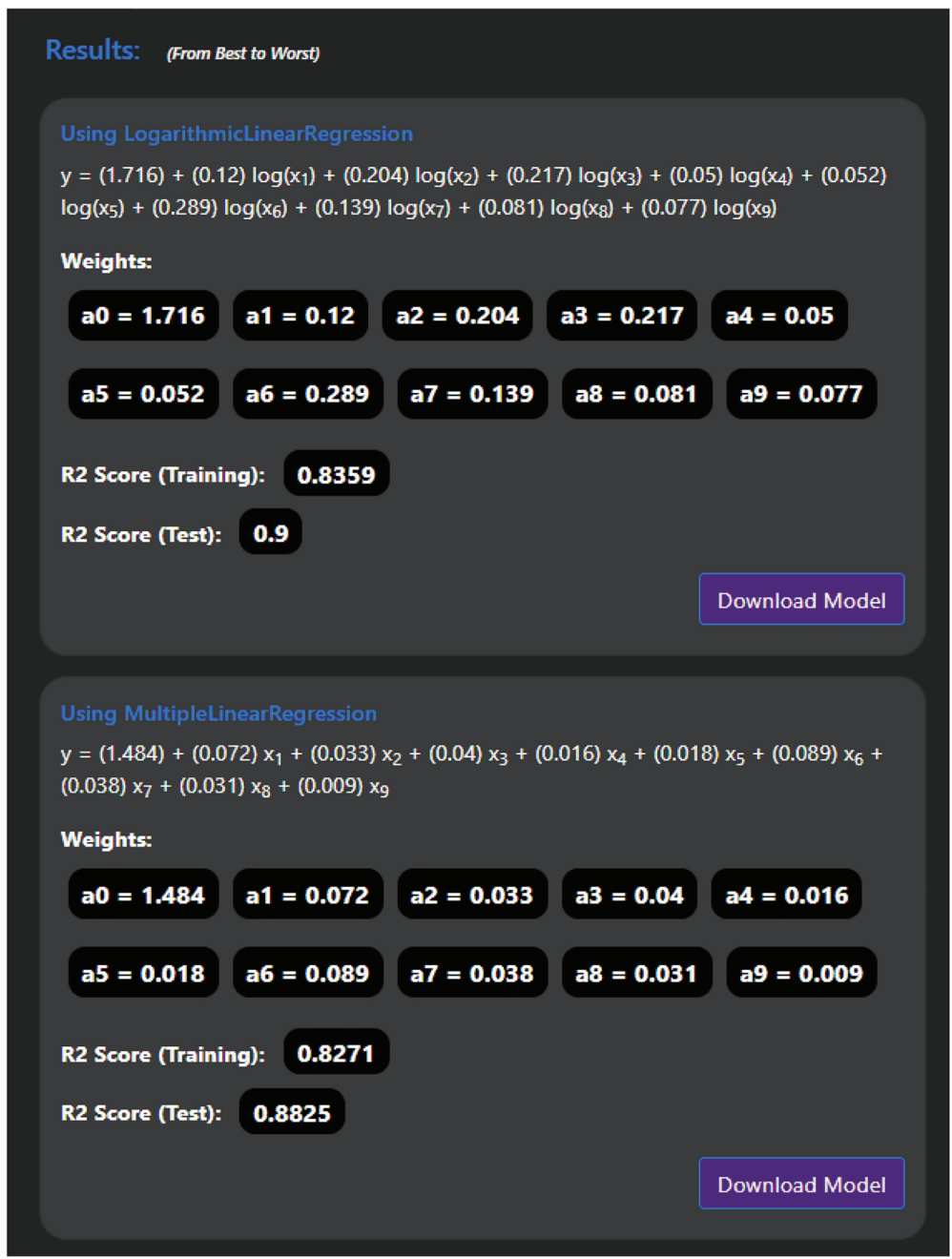

**FIGURE 8**  The resulting regression models sorted from best to worst.

The second open source dataset used to evaluate the system is named "Bike Sharing Data Set."[29] The dataset has features or attributes "instant" (record index), "dteday" (date), "season," "yr" (year), "mnth" (month), "hr" (hour), "holiday" (whether day is holiday or not), "weekday" (day of the week), "workingday" (if day is neither weekend nor holiday is 1, otherwise is 0), "weathersit" (weather situation), "temp" (Normalized temperature in Celsius), "atemp" (Normalized feeling temperature in Celsius), "hum" (Normalized humidity), "windspeed" (Normalized wind speed), "casual" (count of casual users), "registered" (count of registered users), "cnt" (count of total rental bikes including both casual and registered). The "instant" and "dteday" attributes were ignored for training purpose because these two attributes are not correlated with the "cnt" attribute which is the attribute to be predicted through the web tool. The goal here is to use all these feature attributes to predict or explain the variation of the "cnt" attribute which denotes the count of total rental bikes including both casual and registered.

At first, all attributes except "instant," "dteday," "casual," and "registered" were used for regression analysis by the web tool. The resulting best fit model was multiple linear regression model with $r^2$ score of 0.39. This means that approximately 40% variation of the dependent variable "cnt" can be explained by all attributes except "instant," "dteday," "casual," and "registered." The result is shown in Figure 9.

Next "casual" (count of casual users) attribute was considered along with other feature attributes to train the model and to predict the dependent variable "cnt." Considering "casual" attribute along with other attributes in the model training raises the $r^2$ score from 0.39 to 0.62. So it improved multiple linear regression model prediction and also along with other features explained approximately 62% variation of the dependent variable "cnt." The result is shown in Figure 10.

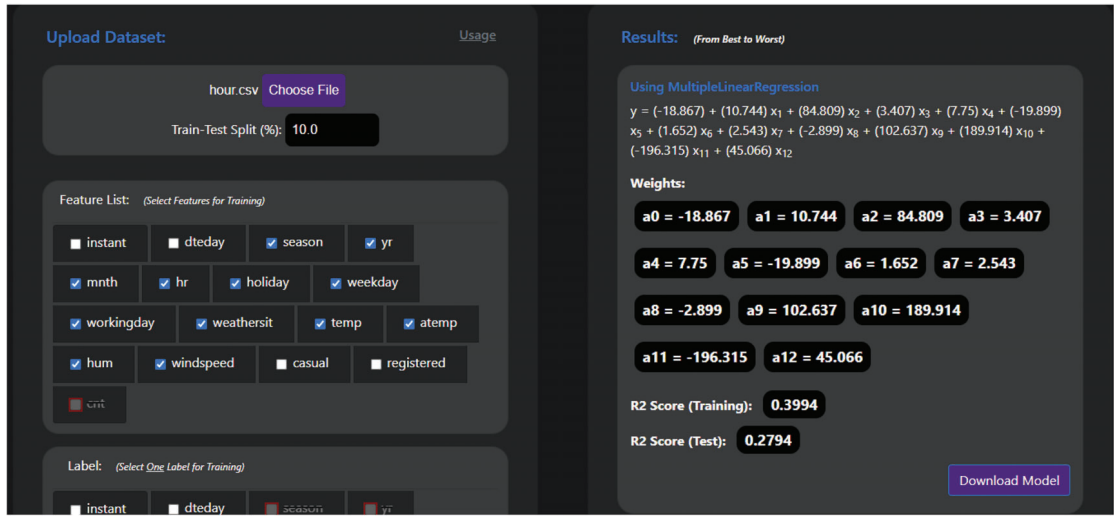

**FIGURE 9** Predicting "cnt" attribute without "casual" and "registered" feature attributes.

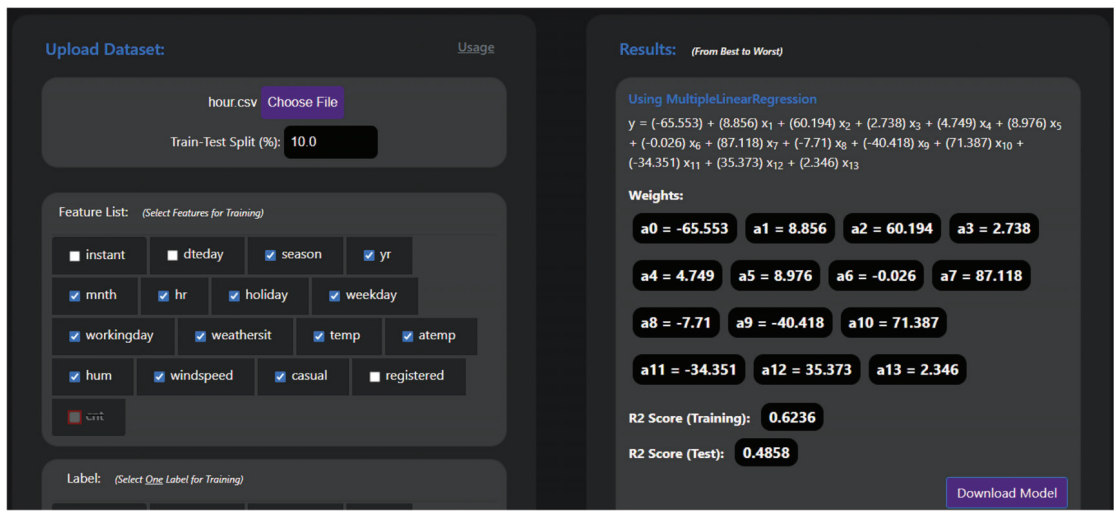

**FIGURE 10** Predicting "cnt" attribute with "casual" attribute along with other attributes except "registered" attribute.

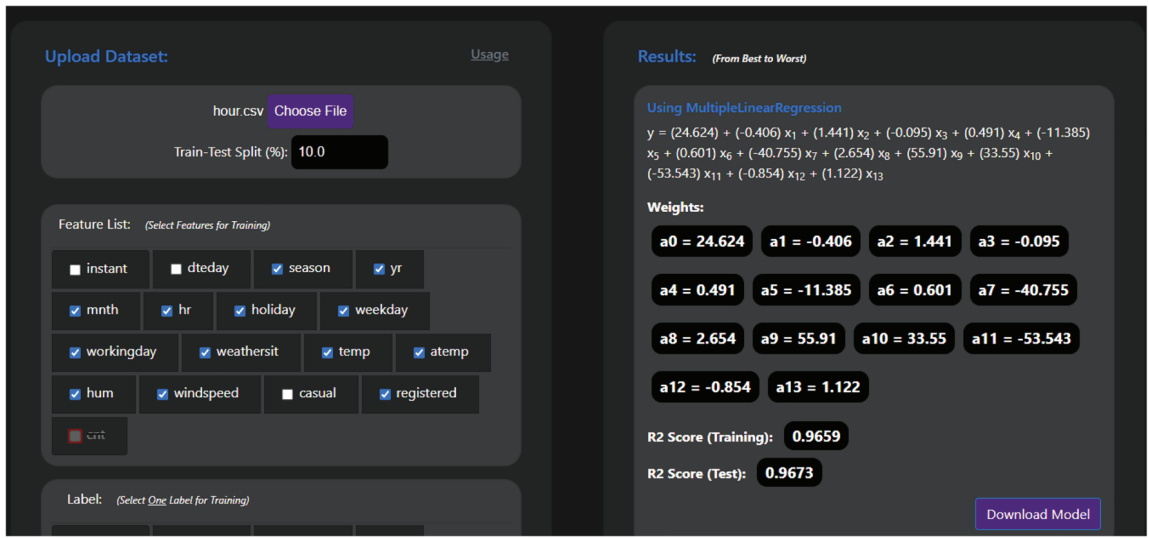

**FIGURE 11** Predicting "cnt" attribute with "registered" attribute along with other attributes except "casual" attribute.

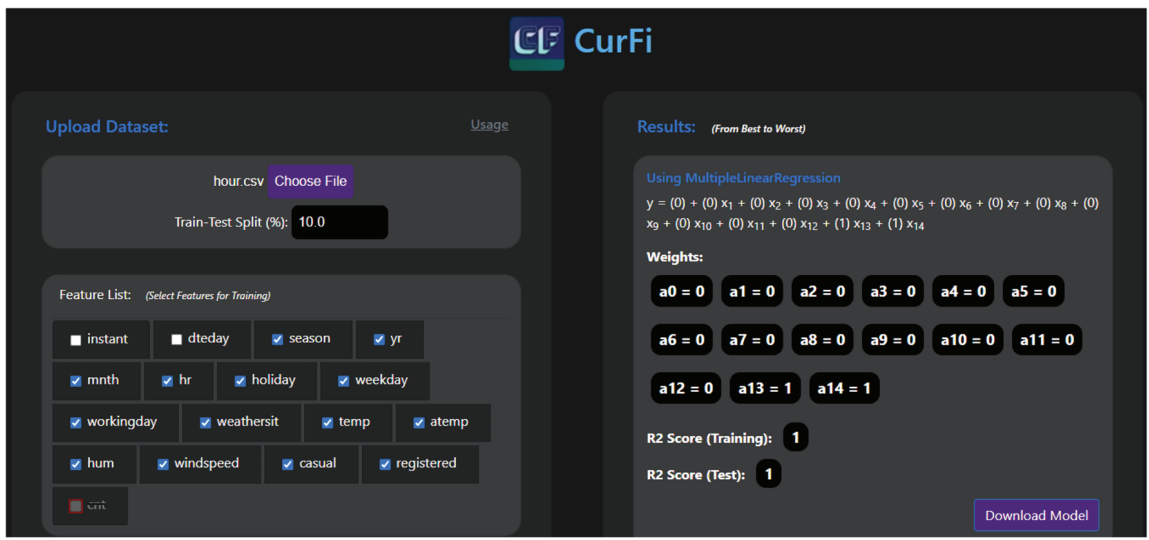

**FIGURE 12** Predicting "cnt" attribute with both "registered" and "casual" attributes along with other attributes.

Next "casual" (count of casual users) attribute was removed from the features list and "registered" (count of registered users) was considered along with other attributes. Considering "registered" attribute hugely improved the model prediction from 0.39 to 0.97. So along with other features, "registered" attribute explained approximately 97% variation of the dependent variable "cnt." The result is shown in Figure 11.

And finally, both "casual" (count of casual users) and "registred" (count of registered users) attributes were considered along with other attributes to train the model. In this case, considering both "casual" and "registered" attributes explained 100% variance of the dependent variable "cnt." In fact the multiple linear regression model in this case indicates that "casual" and "registered" attributes alone can explain 100% variance of the dependent variable "cnt." The result is shown in Figure 12.

In summary, "Bike Sharing Data Set"[29] is another open source dataset that was used as a demo of a real world dataset to test the predictions made by the web tool. This dataset was collected from a real bike sharing system and it contains several feature variables and one dependent variable "cnt" which is the count of total rental bikes including both casual and registered. It is shown that considering different feature attributes affects the overall prediction of the multiple linear regression model which is the best fit model returned by the web app for this particular dataset. This example shows how user can use this web app to try out various combination of feature variables to find out which combination explains the dependent variable best.

**TABLE 2** Summary of finding patterns in open source datasets

| Dataset | Attributes considered | Best fit model returned by web app | Training R2 score |
| --- | --- | --- | --- |
| Breast Cancer Wisconsin[28] | All feature attributes except "Sample code number" | Logarithmic linear regression model | 0.84 |
| Bike Sharing Data Set[29] | All feature attributes except "instant," "dteday," "casual," and "registered" | Multiple linear regression model | 0.39 |
| Bike Sharing Data Set[29] | All feature attributes except "instant," "dteday," and "registered" | Multiple linear regression model | 0.62 |
| Bike Sharing Data Set[29] | All feature attributes except "instant," "dteday," and "casual" | Multiple linear regression model | 0.97 |
| Bike Sharing Data Set[29] | All feature attributes except "instant" and "dteday" | Multiple linear regression model | 1.00 |

Two open source datasets "Breast Cancer Wisconsin"[28] and "Bike Sharing Data Set"[29] were used to evaluate the system. Table 2 summarizes the predictions made by the web application for these two open source datasets.

## 5 | DISCUSSIONS AND CONCLUSIONS

Finding relationships among data points is a very important task for researchers in various fields. For this, they use regression analysis in many cases. But this process is repetitive and time-consuming. In this article the process of automating regression analysis is focused. The developed system discussed in this article finds the relationship for a given dataset automatically by finding the "best fit" model. The system finds the "best fit" model by fitting regression model curves to the dataset. The best curve with the lowest residual error and highest $r^2$ score represents the "best fit" model.

The developed system shows the six linear regression models in the sorted order based on the $r^2$ score. For each model, the system shows the equation representing the relationship between the independent variable(s) and the dependent variable. The system also shows a graphical plot of the data points and the curve representing that model. In this way, the researchers can find and visualize the relationship in the dataset automatically and without any difficulty from the developed system.

Researchers use various existing math softwares and statistical analysis softwares to conduct regression analysis. Also best fit regression models can be derived using normal statistical techniques. For example Fumo et al.[17] used "Mathcad" and "Excel Data Analysis package" software to conduct regression analysis in order to model residential energy consumption. These softwares provide mathematical functions which have to be manually written and run to get the desired result. Olaniyi et al.[18] predicted stock price using linear regression analysis where researchers implemented linear regression models and incorporated these models in a data mining tool. Bradshaw et al.[19] developed a multiple linear regression model to predict maximum oxygen uptake of a person before intense exercise which helps to determine the condition of the cardiorespiratory system of humans. To conduct this regression analysis the researchers used normal statistical techniques. Heim et al.[20] used a multiple linear regression model to predict neuroendocrine stress response in adult women which is an indicator of childhood trauma and major life events. For this the researchers used a famous statistical analysis software SPSS. This tool provides functionality to conduct linear regression analysis efficiently but to use this tool and to interpret the output generated by this tool one must have significant knowledge about linear regression analysis. Chandrashekar Murthy et al.[21] implemented the multiple linear regression model using python and integrated the model in a progressive web app to predict the water demand for domestic purpose of a house. Barhmi et al.[22] developed multiple linear regression model to forecast wind speed manually with normal statistical calculations. Patel et al.[23] predicted the surface roughness level of a machined surface using linear and nonlinear regression model. An automated regression tool would have made the regression analysis much easier to perform. Nyarko-Boateng et al.[24] used a simple linear regression model to predict the location of the fault in the underground optical network. Python sci-kit learn library was used to build the simple linear regression model which requires substantial knowledge of python programming and machine learning models. In summary, all of these math softwares and statistical analysis tools require significant background knowledge about regression analysis and often these tools have their own functions and specifications which the researchers must learn before using them. Moreover, normal statistical techniques for finding the best fit model require significant mathematical calculations which is time-consuming and tedious. Machine learning libraries such as sci-kit learn library also require enough programming skill. But the automated system described in this article alleviates these problems by fully automating the process of conducting linear regression analysis. The output is shown in sorted order in this system and the regression equations are also shown along with graphical plots to assist the users to understand the patterns in the dataset without significant background knowledge.

The system can perfectly fit datasets that have only numerical variables. But one limitation is that datasets with both numerical and categorical variables can not be trained directly but the categorical variables can easily be converted to numerical variables using different encoding techniques. So if a user wishes to train a dataset with one or more categorical variables, then encoding techniques can be used to convert the categorical variables to numerical ones. After that, the user can successfully fit the dataset with the developed system. Support for categorical variables can be added in the future so that users don't have to manually encode the categorical variables into numerical ones. Also the system works well for the datasets that have a linear relationship between independent variables and dependent variable. Since the system does not support any nonlinear regression model or any other machine learning algorithm, it will not be able to produce a good fit for a dataset which has highly nonlinear relationship between independent variables and dependent variable. The system contains only the six linear regression models that are frequently used by researchers. More regression models and machine learning algorithms can be added to the system in the future to make the system more powerful and robust. Also the process of adding new regression models can be automated in the future so that the users can apply new regression models on the dataset as per their requirements. This feature will greatly expand the capability of the system since the users will not be limited to the regression models provided by the system.

## PEER REVIEW

The peer review history for this article is available at https://publons.com/publon/10.1002/eng2.12522.

## DATA AVAILABILITY STATEMENT

Data sharing is not applicable to this article as no new data were created or analyzed in this study.

## CONFLICT OF INTEREST

The authors declare no potential conflict of interest

## AUTHOR CONTRIBUTIONS

**Ayon Roy:** conceptualization (equal); formal analysis (equal); software (equal); validation (equal); writing – original draft (equal); writing – review and editing (equal). **Tausif Al Zubayer:** conceptualization (equal); software (equal); writing – original draft (equal).

## ORCID

*Muhammad Nazrul Islam*  https://orcid.org/0000-0002-7189-4879

## REFERENCES

1. Gulsen M, Smith AE, Tate DM. A genetic algorithm approach to curve fitting. *Int J Product Res*. 1995;33(7):1911-1923. doi:10.1080/00207549508904789
2. Ledvij M. Curve fitting made easy. *Ind Phys*. 2003;9(2):24-27.
3. Ranganathan S, Nakai K, Schonbach C. *Encyclopedia of Bioinformatics and Computational Biology: ABC of Bioinformatics*. 1st ed. Elsevier Science Publishers B. V., NLD; 2018.
4. Li Y, Arce GR. A maximum likelihood approach to least absolute deviation regression. *EURASIP J Adv Signal Process*. 2004;2004(12):1-8.
5. Schlossmacher EJ. An iterative technique for absolute deviations curve fitting. *Journal of the American Statistical Association*. 1973;68(344):857-859.
6. Bellman R, Roth R. Curve fitting by segmented straight lines. *J Am Stat Assoc*. 1969;64(327):1079-1084. doi:10.1080/01621459.1969.10501038
7. Uyanık GK, Güler N. A study on multiple linear regression analysis. *Proc Soc Behav Sci*. 2013;106:234-240. doi:10.1016/j.sbspro.2013.12.027
8. Palmer P, O'Connell D. Regression analysis for prediction: understanding the process. *Cardiopulm Phys Ther J*. 2009;20:23-26.

9. Leatherbarrow RJ. Using linear and non-linear regression to fit biochemical data. *Trends Biochem Sci*. 1990;15(12):455-458. doi:10.1016/0968‐0004(90)90295‐m
10. Andrews DF. A robust method for multiple linear regression. *Technometrics*. 1974;16(4):523-531. doi:10.1080/00401706.1974.10489233
11. Poole M, O'Farrell P. The assumptions of the linear regression model. *Trans Inst Br Geograph*. 1971;52:145-158. doi:10.2307/621706
12. Isobe T, Feigelson ED, Akritas MG, Babu GJ. Linear regression in astronomy. *Astrophys J*. 1990;364:104. doi:10.1086/169390
13. Dabhi JA. A web-app for creating machine learning models interactively; 2021.
14. Nokeri TC. Integrating a machine learning algorithm into a web app. *Web App Development and Real-Time Web Analytics with Python*. Apress; 2021. doi:10.1007/978‐1‐4842‐7783‐6&uscore;11
15. Singh P. Machine learning deployment as a web service. *Deploy Machine Learning Models to Production*. Apress; 2021. doi:10.1007/978‐1‐4842‐6546‐8&uscore;3
16. Roy A. Curfi.js. NPM package; 2021. https://www.npmjs.com/package/curfi
17. Fumo N, Biswas M. Regression analysis for prediction of residential energy consumption. *Renew Sustain Energy Rev*. 2015;47:332-343. doi:10.1016/j.rser.2015.03.035
18. Olaniyi SAS, Adewole S, Jimoh R. Stock trend prediction using regression analysis-A data mining approach. *ARPN J Syst Softw*. 2011;1:154-157.
19. Bradshaw D, George J, Hyde A, et al. An accurate VO2 max nonexercise regression model for 18–65-year-old adults. *Res Quart Exercise Sport*. 2005;76:426-432. doi:10.1080/02701367.2005.10599315
20. Heim C, Newport DJ, Wagner D, Wilcox MM, Miller AH, Nemeroff CB. The role of early adverse experience and adulthood stress in the prediction of neuroendocrine stress reactivity in women: a multiple regression analysis. *Depress Anxiety*. 2002;15(3):117-125. doi:10.1002/da.10015 PMID: 12001180.
21. Chandrashekar Murthy BN, Balachandra HN, Sanjay NK, Chakradhar Reddy C. Prediction of water demand for domestic purpose using multiple linear regression. In: Smys S, Iliyasu AM, Bestak R, Shi F, eds. *New Trends in Computational Vision and Bio-inspired Computing. ICCVBIC*. Springer; 2020. doi:10.1007/978‐3‐030‐41862‐5&uscore;81
22. Barhmi S, Elfatni O, Belhaj I. Forecasting of wind speed using multiple linear regression and artificial neural networks. *Energy Syst*. 2020;11:935-946. doi:10.1007/s12667‐019‐00338‐y
23. Patel DR, Kiran MB, Vakharia V. Modeling and prediction of surface roughness using multiple regressions: a noncontact approach. *Eng Rep*. 2020;2:e12119. doi:10.1002/eng2.12119
24. Nyarko-Boateng O, Adekoya AF, Weyori BA. Predicting the actual location of faults in underground optical networks using linear regression. *Eng Rep*. 2021;3:e12304. doi:10.1002/eng2.12304
25. Marill KA. Advanced statistics: linear regression, Part II: multiple linear regression. *Acad Emerg Med*. 2008;11:94-102. doi:10.1197/j.aem.2003.09.006
26. Chapra SC, Canale RP. *Numerical Methods for Engineers*. McGraw-Hill Higher Education; 2006.
27. Roy A. CurFi; 2021. https://curfi.netlify.app/
28. Blake C. UCI repository of machine learning databases; 1998. https://archive.ics.uci.edu/ml/datasets/Breast+Cancer+Wisconsin+%28Diagnostic%29
29. Fanaee-T H, Gama J. Event labeling combining ensemble detectors and background knowledge. *Progress in Artificial Intelligence*. Springer; 2013:1-15. https://archive.ics.uci.edu/ml/datasets/Bike+Sharing+Dataset